\documentclass[conference]{IEEEtran}
\IEEEoverridecommandlockouts
\usepackage{cite}
\usepackage{amsmath,amssymb,amsfonts}
\usepackage{algorithmic}
\usepackage{graphicx}
\usepackage{textcomp}
\usepackage{csquotes}
\usepackage{xcolor}
\usepackage{adjustbox}
\usepackage{hyperref}
\usepackage{multirow}
\def\BibTeX{{\rm B\kern-.05em{\sc i\kern-.025em b}\kern-.08em
    T\kern-.1667em\lower.7ex\hbox{E}\kern-.125emX}}
\DeclareMathOperator{\argmin}{argmin}
\begin{document}

\title{LiBeamsNet: AUV Velocity Vector Estimation in Situations of Limited DVL Beam Measurements}

\author{\IEEEauthorblockN{Nadav Cohen\IEEEauthorrefmark{1} and Itzik Klein}
\IEEEauthorblockA{\textit{The Hatter Department of Marine Technologies} \\
\textit{Charney School of Marine Sciences, University of Haifa}\\
Haifa, Israel}

\thanks{\IEEEauthorrefmark{1}Corresponding author: N. Cohen (email: ncohe140@campus.haifa.ac.il).}}

\maketitle

\begin{abstract}
Autonomous underwater vehicles (AUVs) are employed for marine applications and can operate in deep underwater environments beyond human reach. A standard solution for the autonomous navigation problem can be obtained by fusing the inertial navigation system and the Doppler velocity log sensor (DVL). The latter measures four beam velocities to estimate the vehicle's velocity vector. In real-world scenarios, the DVL may receive less than three beam velocities if the AUV operates in complex underwater environments. In such conditions, the vehicle's velocity vector could not be estimated leading to a navigation solution drift and in some situations the AUV is required to abort the mission and return to the surface. To circumvent such a situation, in this paper we propose a deep learning framework, LiBeamsNet, that utilizes the inertial data and the partial beam velocities to regress the missing beams in two missing beams scenarios. Once all the beams are obtained, the vehicle's velocity vector can be estimated. The approach performance was validated by sea experiments in the Mediterranean Sea. The results show up to 7.2 \% speed error in the vehicle's velocity vector estimation in a scenario that otherwise could not provide an estimate.
\end{abstract}


\section{Introduction}
The primary purpose of the autonomous underwater vehicle (AUV) is to perform scientific tasks in the great depths of the ocean. In order to do so, the AUV has to operate autonomously in the underwater environment. Thus, it contains several sensors collecting data to enable autonomous navigation \cite{jain2015review}.\\
The inertial navigation system (INS) provides the navigation solution utilizing inertial measurements from its three-axis accelerometer and a three-axis gyroscope; providing the specific force and angular velocity vectors, respectively \cite{titterton2004strapdown,groves2015principles}. The navigation solution gives the AUV's position, velocity, and orientation. Although this information is sufficient for autonomous navigation, the INS cannot be used as a standalone solution due to its nature to accumulate error over time \cite{farrell2007gnss,shin2002accuracy}.
To that end, a Doppler velocity log (DVL) is typically used in the AUV to overcome this problem. \\
The DVL is based on the Doppler effect, while the bottom-lock refers to a situation where four beams are transacted to the seafloor and reflected back to the sensor. The DVL can achieve a typical velocity measurement accuracy of 0.2\% of the current velocity and is, therefore, considered an accurate sensor \cite{wang2019novel}. In most situations, an accurate navigation solution can be obtained by performing a fusion between the INS and the DVL for error accumulation avoidance \cite{yonggang2013tightly,yao2020simple,klein2015observability}. However, if the process noise covariance of the filter is not determined properly, the navigation solution may drift. To circumvent such situations both model and learning adaptive filter solution exists in the literature \cite{or2022hybrid,or2022Adaptive}. 
As the DVL is the aiding sensor to the navigation solution, it is crucial to obtain continuous DVL data flow. However, in real-world scenarios, the DVL may receive only partial beam measurements in different situations. These situations include sea creatures blocking the acoustics beams, trenches on the seafloor, and extreme pitch and roll maneuvers (such as diving) \cite{eliav2018ins,klein2020continuous}. The minimum amount of measured beams for obtaining an AUV velocity estimate is three. If there are fewer, an AUV velocity vector cannot be derived, which results in an error accumulation of the navigation solution. Usually, in this situation, the AUV will abort the mission and will be forced to surface.\\
To cope with such situations, several solutions were suggested in the literature. In \cite{tal2017inertial}  additional data and the current partial beams measurements were employed to create virtual beams enabling the estimation of the AUV velocity. In \cite{liu2018ins}, the partial beams were used directly in the INS/DVL sensor fusion in a tightly coupled (TC) approach. Furthermore, a data-driven method was used to compensate for a situation of one missing beam using past DVL measurements and Long short-term memory (LSTM) model and outperformed the model-based approach \cite{yona2021compensating}. Recently, the INS/DVL performance using a tightly coupled approach in
situations of partial DVL beam measurement, was improved
using a learning virtual beam aided solution \cite{yao2022Virtual}. \\
In situations of complete DVL measurement, we proposed a deep learning approach, named BeamsNet, to estimate the AUV velocity vector and replace the model-based approach \cite{cohen2022beamsnet}. We demonstrated our approach's ability to outperform the model-based approach using a dataset collected from sea experiments made with an AUV.
Leveraging from that work, in this paper, a deep learning approach, LiBeamsNet (LImited Beams NETwork), is proposed to compensate for a situation of two missing DVL beams. Our network uses inertial readings and the current partial beams as input and provides the two missing beams. Once all the beams are obtained, they can be used to estimate the AUV velocity vector, which otherwise would not be available. To evaluate the suggested approach performance, sea experiments were conducted in the Mediterranean Sea with the University of Haifa's Snapir AUV. Approximately four and a half hours of recorded accelerometers, gyroscopes, and DVL data was collected and used to train, validate, and test the network.\\
The rest of the paper is organized as follows: Section \ref{DVL} describes the DVL equations and error models. Section \ref{data} introduces the proposed approach for compensating for the missing beams and the network architecture. In Section \ref{AandR}, an analysis of the results is shown, and the conclusions are discussed in Section \ref{con}. 
\section{DVL Velocity Calculations}\label{DVL}
\begin{figure}[h]
	\centering
		\includegraphics[scale=.30,]{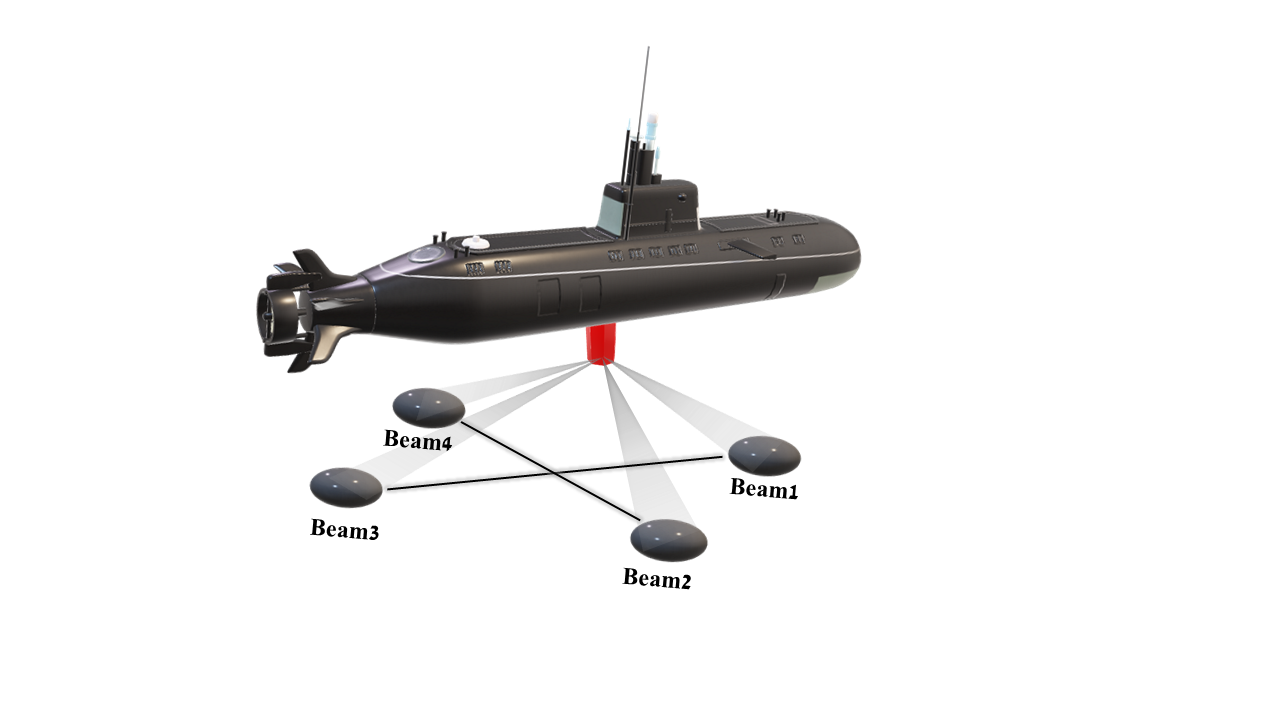}
	  \caption{DVL transmits acoustic beams in the "Janus Doppler configuration".}\label{fig1}
\end{figure}

The purpose of the DVL sensor is to supply the AUV velocity vector. To do so, the DVL transmits four acoustic beams in different directions to the seafloor, and once they are reflected back to the sensor, the beam velocity is obtained due to the frequency shift. The DVL's transducers are commonly configured in a \enquote*{$\times$} shape configuration, known in the literature as the "Janus Doppler configuration" \cite{brokloff1994matrix}, as shown in Figure \ref{fig1}. 
\\
The beam velocity measurements are the AUV velocity vector projected to the DVL's transactors' direction. Therefore, by looking at the geometry of the problem, the direction of each beam in the DVL's body frame can be expressed as \cite{liu2018ins}:  
\begin{equation}\label{eqn:1}
    \centering
        \boldsymbol{b}_{\dot{\imath}}=
        \begin{bmatrix} 
        \cos{\psi_{\dot{\imath}}}\sin{\theta}\quad
        \sin{\psi_{\dot{\imath}}}\sin{\theta}\quad
        \cos{\theta}
    \end{bmatrix}_{1\times3}
\end{equation} 
where $\dot{\imath}=1,2,3,4$ represents the beam number and $\psi$ and $\theta$ are the heading and pitch angles relative to the body frame, respectively \cite{tal2017inertial}. The pitch angle is fixed, and the heading angle can be expressed by \cite{yona2021compensating}: 
\begin{equation}\label{eqn:2}
    \centering
        \psi_{\dot{\imath}}=(\dot{\imath}-1)\cdot\frac{\pi}{2}+\frac{\pi}{4}\;[rad]\;,\; \dot{\imath}=1,2,3,4
\end{equation}
 hence, the relation between the AUV velocity in body frame, $\boldsymbol{v}_{b}^{b}$, to the beam velocity measurements, $\boldsymbol{\upsilon}_{beam}$, can be written, using the transformation matrix $\mathbf{H}$, as:
\begin{equation}\label{eqn:3}
    \centering
        \boldsymbol{\upsilon}_{beam}=\mathbf{H}\boldsymbol{v}_{b}^{b} ,\quad
        \mathbf{H}=
        \begin{bmatrix} 
            \boldsymbol{b}_{1}\\\boldsymbol{b}_{2}\\\boldsymbol{b}_{3}\\\boldsymbol{b}_{4}\\
    \end{bmatrix}_{4\times3}
\end{equation} 
Due to limitations of the sensor, the DVL is subject to errors, and therefore a beam error model is derived. As a consequence, a bias, scale factor, and zero-mean Gaussian processing noise are added to the beam velocity measurements from \eqref{eqn:3}, yielding
\begin{equation}\label{eqn:4}
    \centering
        \boldsymbol{y}= \mathbf{H}(\boldsymbol{v}_{b}^{b}\cdot(1+\boldsymbol{s}_{DVL}))+\boldsymbol{b}_{DVL}+\boldsymbol{n}
\end{equation}
where \begin{itemize}
    \item $\boldsymbol{b}_{DVL}$ is the $4\times1$ bias vector
    \item $\boldsymbol{s}_{DVL}$ is the $4\times1$ scale factor vector
    \item $\boldsymbol{n}$ is the zero white mean Gaussian processing noise
    \item $\boldsymbol{y}$ is the beam velocity measurements
\end{itemize}
In order to get a good estimate of the AUV velocity vector, the beam measurements need to be filtered and transformed. To that end, a linear least squares (LS) filter is used:
\begin{equation}\label{eqn:5}
    \centering
        \hat{\boldsymbol{v}}_{b}^{b}=
        \underset{\boldsymbol{v}_{b}^{b}}{\argmin}{\mid\mid\boldsymbol{y}-\mathbf{H}\boldsymbol{v}_{b}^{b} \mid\mid}^{2}
\end{equation} 
The solution for this filter is the pseudo inverse of matrix $\mathbf{H}$ times the beam's velocity measurement $\boldsymbol{y}$, and it does the two operations. It filters the additive noise and bias and transforms the vector from the beams coordinate frame to the AUV body frame. The solution for this estimator $\hat{\boldsymbol{v}}_{b}^{b}$ can be seen below \cite{braginsky2020correction}:
\begin{equation}\label{eqn:6}
    \centering
        \hat{\boldsymbol{v}}_{b}^{b}=(\mathbf{H}^{T}\mathbf{H})^{-1}\mathbf{H}^{T}\boldsymbol{y}
\end{equation} 
\section{Compensating for Situations of Limited  Beam Measurements}\label{data}
\begin{figure}[h!]
	\centering
		\includegraphics[scale=.30,]{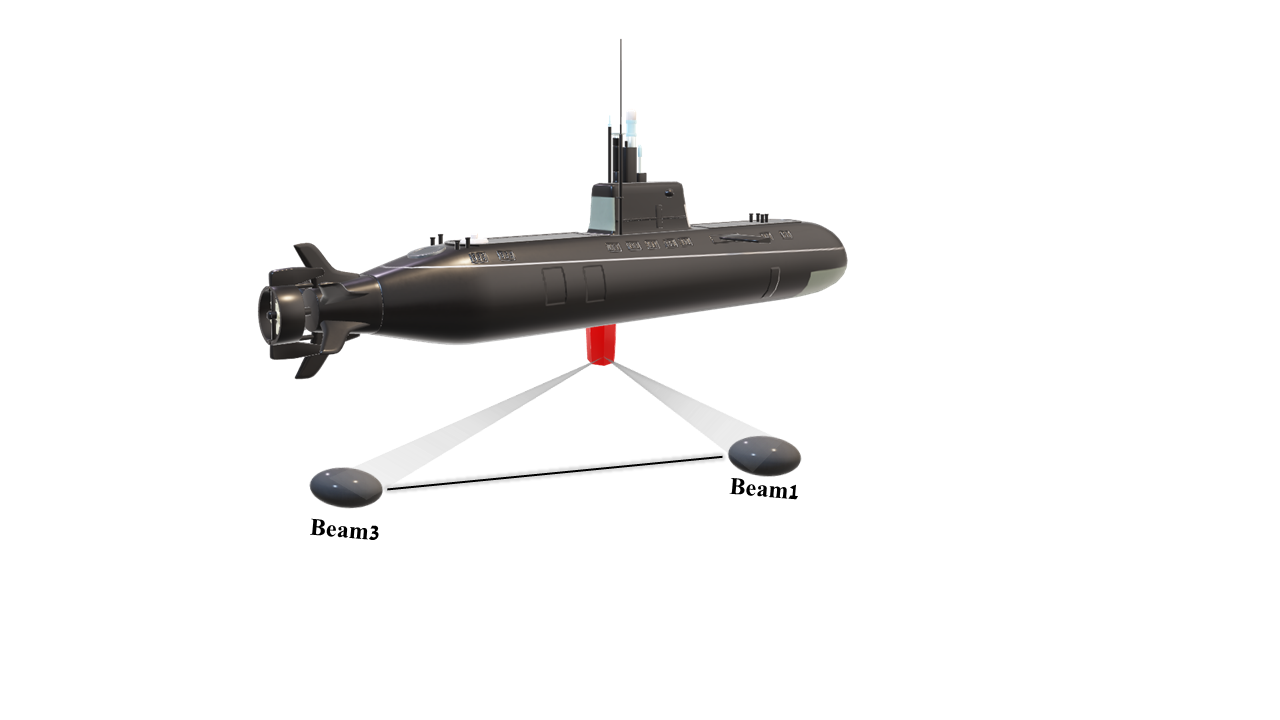}
	  \caption{A scenario with only two available DVL beams.}\label{fig2}
\end{figure}

\begin{figure}[h!]
	\centering
		\includegraphics[width=1\columnwidth]{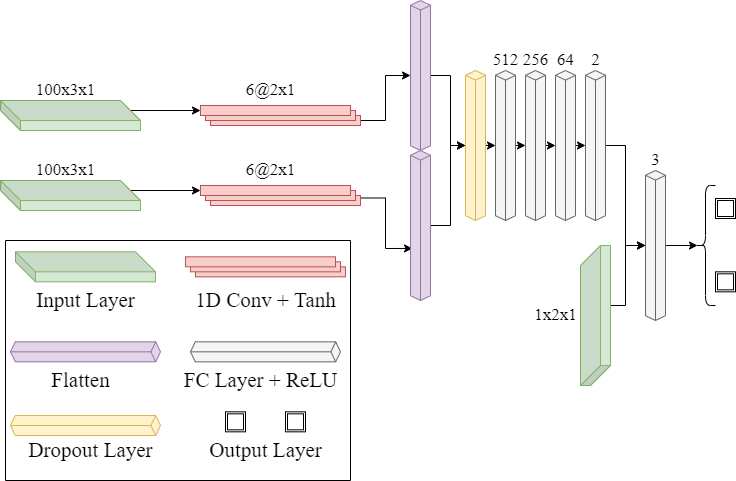}
	  \caption{LiBeamsNet network architecture.}\label{fig4}
\end{figure}

\begin{figure*}[h!]
	\centering
		\includegraphics[width=0.8\textwidth]{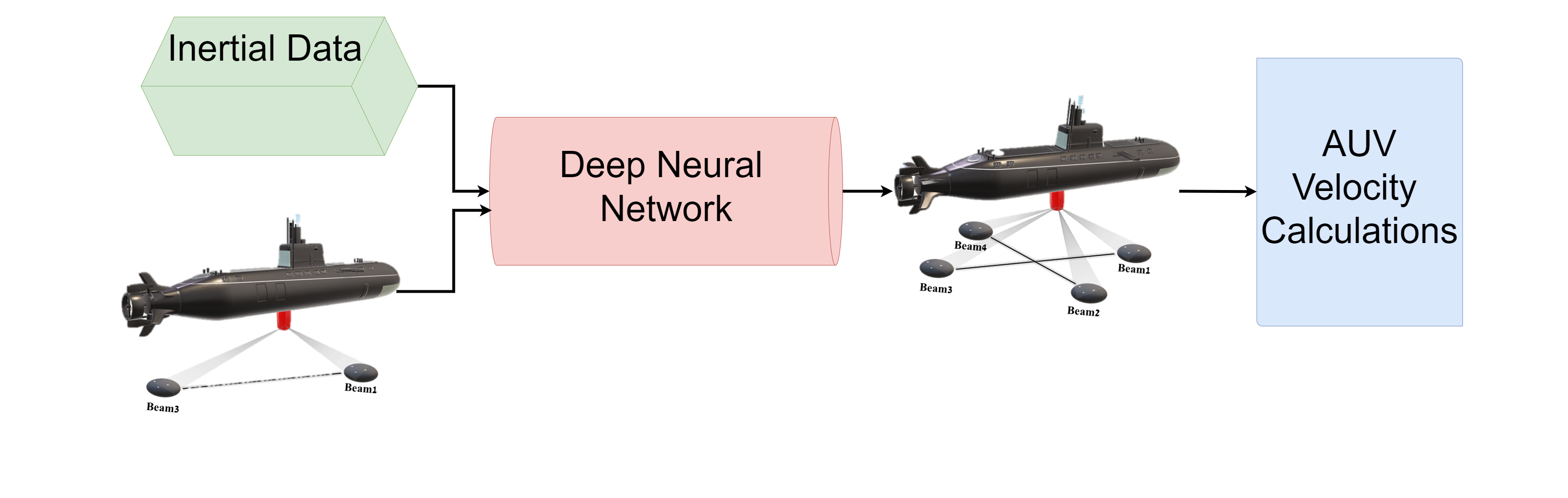}
	  \caption{A block diagram showing the flow of information in LiBeamsNet framework.}\label{fig3}
\end{figure*}

At least three beam measurements are required to estimate the AUV velocity vector using (6) \cite{brokloff1994matrix}. That, in a cost of accuracy, the DVL algorithm can compensate for one missing beam. Here, we focus on the case of two missing beams. In such scenarios the velocity solution by (6) cannot be calculated. Specifically, we assume beams two and four are not available,  as can be seen in Figure \ref{fig2}. However, one can easily adjust our approach to handle any combination of two missing beams.
\\
In \cite{cohen2022beamsnet},BeamsNet framework showed promising results in situations of complete DVL measurements. Therefore, we utilize the same network structure with several modifications to handle missing beams scenarios and present LiBeamsNet. To that end, instead of using all four beams as input to the network, now only the current partial beam measurements are required together with the inertial sensor data (specific force and angular velocity measurements). In addition, the current network regresses the two missing beams instead of regressing the AUV velocity vector directly as in BeamsNet. LiBeamsNet network architecture is presented in Figure~\ref{fig4}. The inertial data from the accelerometer and gyroscope passes through a one-dimensional convolutional layer consisting of six filters of size $2 \times 1$. By doing so, features are extracted from the data and then flattened, combined, and passed through a dropout layer with $p = 0.2$. After a sequence of fully connected layers, the current partial beam measurements are combined and moved through the last fully connected layer that produces the $2 \times 1$ vector, which is the regressed missing beams. The architecture and the activation functions after each layer are presented in figure \ref{fig4}. The network trained over 50 epochs, mini-batch size of 32, 0.01 learning rate, and RMSprop optimizer.
Once all four required beams are obtained (two measured and two regressed using LiBeamsNet), the AUV velocity calculations can be made using the model-based approach or using BeamsNet. In this paper, we use the model base approach. A block diagram of our method can be seen in Figure~\ref{fig3}. 
\section{Analysis and Results}\label{AandR}
\begin{figure}[h!]
	\centering
		\includegraphics[width=0.9\columnwidth]{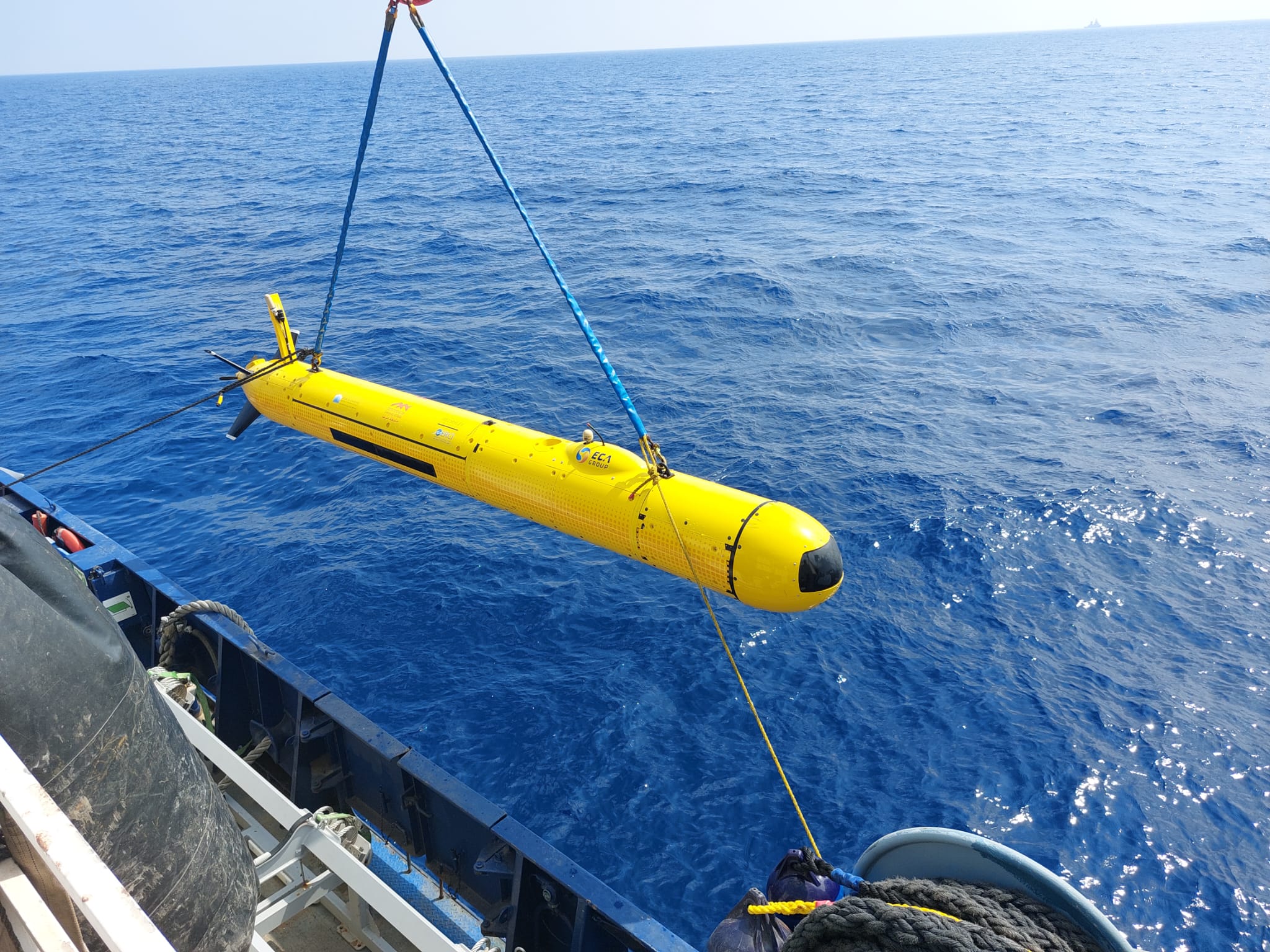}
	  \caption{"Snapir" AUV being dropped into a mission in the Mediterranean Sea.}\label{fig5}
\end{figure}

Our LiBeamsNet framework was validated using sea experiment data recorded by the "Snapir" AUV, which is A18-D, ECA GROUP mid-size AUV. "Snapir",presented in Figure~\ref{fig5}, has a length of 5.5 [m], a diameter of 0.5 [m], 24 hours’ endurance, and a depth rating of 3000 [m]. "Snapir" is equipped with Teledyne RD Instruments, Navigator DVL \cite{Teledyne}, and iXblue Phins Subsea IMU \cite{iXblue}. The train and validation sets were created by collecting inertial data, and DVL data from nine different missions in the Mediterranean Sea, varied in their length, maneuvers, and depth, performed by the AUV with a total time duration of 13,886 seconds. Since the DVL samples were taken at 1Hz and the inertial data at 100Hz,  13,886 and 1,388,600 measurements, respectively, are available in the dataset. An additional dataset was created using "Snapir" AUV on a different day with different sea conditions. This dataset is used as our test set to examine the generalization of our proposed approach. It contains 2001 DVL measurements and 200,100 IMU measurements. All of the data can be found on \url{https://github.com/ansfl/BeamsNet}.\\
In an ideal situation, the AUV will carry two DVL sensors. One would be the ground truth, and the other the unit under test. Since it was not the case, to create the unit under test DVL data, the error model,  \eqref{eqn:4} was implemented with scale factor, bias, and STD of the zero-mean white Gaussian noise of 0.7\%, 0.0001[m/s], and 0.042[m/s], respectively. Once the data was created, only beams one and three were used to mimic a situation of partial beam measurements.
The matrices which were chosen to evaluate the suggested framework: 1) root mean squared error (RMSE) 2) mean absolute error (MAE) 3) the coefficient of determination ($R^{2}$), and 4) the variance account for (VAF). The RMSE and MAE express the velocity error in units of $[m/s]$, while the $R^{2}$ and VAF are unitless. Those matrices are defined as follows:
\begin{equation}\label{eqn:7}
    \centering
        RMSE(\boldsymbol{x}_{\dot\imath},\hat{\boldsymbol{x}}_{\dot\imath})=\sqrt{\frac{\sum_{\dot\imath=1}^{N}(\boldsymbol{x}_{\dot\imath}-\hat{\boldsymbol{x}}_{\dot\imath})^{2}}{N}}
\end{equation}
\begin{equation}\label{eqn:8}
    \centering
        MAE(\boldsymbol{x}_{\dot\imath},\hat{\boldsymbol{x}}_{\dot\imath})=\frac{\sum_{\dot\imath=1}^{N}|\boldsymbol{x}_{\dot\imath}-\hat{\boldsymbol{x}}_{\dot\imath}|}{N}
\end{equation}
\begin{equation}\label{eqn:9}
    \centering
        R^{2}(\boldsymbol{x}_{\dot\imath},\hat{\boldsymbol{x}}_{\dot\imath})=1- \frac{\sum_{\dot\imath=1}^{N}(\boldsymbol{x}_{\dot\imath}-\hat{\boldsymbol{x}}_{\dot\imath})^{2}}{\sum_{\dot\imath=1}^{N}(\boldsymbol{x}_{\dot\imath}-\bar{\boldsymbol{x}}_{\dot\imath})^{2}}
\end{equation}
\begin{equation}\label{eqn:10}
    \centering
        VAF(\boldsymbol{x}_{\dot\imath},\hat{\boldsymbol{x}}_{\dot\imath})=[1-\frac{var(\boldsymbol{x}_{\dot\imath}-\hat{\boldsymbol{x}}_{\dot\imath})}{var(\boldsymbol{x}_{\dot\imath})}]\times100
\end{equation}
where N is the number of samples, $\boldsymbol{x}_{\dot\imath}$ is the ground truth velocity vector norm of the DVL, $\hat{\boldsymbol{x}}_{\dot\imath}$ is the predicted velocity vector norm of the AUV, generated after regressing the two missing beams by the network, $\bar{\boldsymbol{x}}_{\dot\imath}$ is the mean of the ground truth velocity vector norm of the DVL, and $var$ stands for variance. Note that if the VAF is $100$, the $R^{2}$ is $1$, and the RMSE and MAE are $0$, the model is considered outstanding \cite{armaghani2021comparative}. In addition, the mean of the ground truth velocity vector norm was calculated to evaluate the magnitude of the speed error. \\
Results for the validation and test datasets are given in Table \ref{Tbl:1}. A speed average of $1.14 [m/s]$ and $1.34 [m/s]$ were computed for the validation and test sets, respectively. These values were used as references to the calculated errors. The results indicate that LiBeamsNet approach performs well in regressing the missing beams and estimating the velocity vector. Looking at the validation set results, we can see that the RMSE and the MAE are close to $0$ and yield 2.5 \% and 1.54 \% speed errors, respectively. The $R^{2}$ and VAF metrics suggest that the proposed deep learning approach has good statistical performance. Furthermore, the suggested approach was evaluated on a test set in order to examine the network's robustness. The network yields 7.2 \% and 3.32 \% speed error with respect to the RMSE and MAE metrics, respectively. The $R^{2}$ and VAF criteria also show good statistical performance. However, the test set did not perform as well as the validation set, indicating overfitting or lack of robustness. 
\begin{table}[h!]
\caption{Estimated AUV speed RMSE, MAE, $R^{2}$, and VAF after the beams compensation using LiBeamsNet on both the validation and test sets}
\centering
   \begin{adjustbox}{width=\columnwidth}
\begin{tabular}{|c|c|c|}
\hline
Evaluation Metrics & \multicolumn{1}{l|}{Validation Set} & \multicolumn{1}{l|}{Test Set} \\ \hline
RMSE $[m/s]$       & 0.03                             & 0.08                       \\ \hline
RMSE $[\%]$        & 2.5                                 & 7.2                           \\ \hline
MAE $[m/s]$        & 0.01                             & 0.03                       \\ \hline
MAE $[\%]$         & 1.54                                & 3.32                          \\ \hline
$R^{2}$            & 0.99                             & 0.98                       \\ \hline
VAF                & 99.73                            & 99.01                      \\ \hline
\end{tabular}
\end{adjustbox}
    \label{Tbl:1}
\end{table}
\section{Conclusions}\label{con}
This paper suggested LiBeamsNet, a deep-learning approach to compensate for a case of two missing beams in the DVL reading using inertial data. Specifically, when beams two and four are missing, and the DVL cannot provide the AUV velocity estimation. The network input is the current beams and the inertial data. The network processes them through a multi-headed 1DCNN neural network, and outputs the two missing beams. Once they were obtained, all four beams were plugged into the model-based approach to estimate the AUV velocity vector that otherwise would not be available. To evaluate the approach, sea experiments were conducted in the Mediterranean Sea using "Snapir" AUV on two different dates and sea conditions to obtain both train and validation sets from one experiment and a test set unfamiliar to the network from the other experiment.\\
The results show that the suggested approach can regress the missing beams with good accuracy, resulting in up to 7.2 \% speed error. Nevertheless, there are differences between the validation and test sets' performance which could indicate overfitting and the need to train over more data.\\
Using our approach, the DVL can compensate for limited beam measurements, estimate the AUV velocity, and allow it to complete its mission. Although evaluated for one scenario of two missing beams (beams 2 and 4), the proposed approach can be applied for any set of two missing beams and can be easily adjusted to handle situations of three missing beams.

\section*{Acknowledgments}
N.C. is supported by the Maurice Hatter Foundation.
\newpage
\bibliographystyle{ieeetr}
\bibliography{refs}

\end{document}